\documentclass[]{interact}

\usepackage{epstopdf}
\usepackage[caption=false]{subfig}

\usepackage[numbers,sort&compress]{natbib}
\bibpunct[, ]{[}{]}{,}{n}{,}{,}

\theoremstyle{plain}

\theoremstyle{definition}

\theoremstyle{remark}

\usepackage[export]{adjustbox}

\usepackage{amsmath}

\begin{document}


\title{Dance of Fireworks: An Interactive Broadcast Gymnastics Training System Based on Pose Estimation}

\author{
\name{Haotian Chen\textsuperscript{a,d}, Ziyu Liu\textsuperscript{b}, Xi Cheng\textsuperscript{c,d} and Chuangqi Li\textsuperscript{*d}\thanks{CONTACT Chuangqi Li. Email: lichuangqi@stu.htu.edu.cn}}
\affil{\textsuperscript{a}SWJTU-LEEDS Joint School, Southwest Jiaotong University, Chengdu, 610031, CN; \textsuperscript{b}School of Computer and Information Engineering, 453007, Xinxiang, Henan, CN; \textsuperscript{c}Xi 'an Jiaotong-Liverpool University, 215123, Suzhou, Jiangsu, CN;\textsuperscript{d}School of Vehicle and Delivery, Tsinghua University, 10084, Beijing, CN;
}
}

\maketitle

\begin{abstract}

This study introduces Dance of Fireworks, an interactive system designed to combat sedentary health risks by enhancing engagement in radio calisthenics. Leveraging mobile device cameras and lightweight pose estimation (PoseNet/TensorFlow Lite), the system extracts body keypoints, computes joint angles, and compares them with standardized motions to deliver real-time corrective feedback. To incentivize participation, it dynamically maps users’ movements (e.g., joint angles, velocity) to customizable fireworks animations, rewarding improved accuracy with richer visual effects. Experiments involving 136 participants demonstrated a significant reduction in average joint angle errors from 21.3° to 9.8° (p<0.01) over four sessions, with 93.4 percent of users affirming its exercise-promoting efficacy and 85.4 percent praising its entertainment value. Distinctively, the system operates without predefined motion templates or specialized hardware, enabling seamless integration into office environments. Future enhancements focus on refining pose recognition (via models like EfficientNet), minimizing latency, and adding features such as multiplayer interaction and music synchronization. By merging computer vision with gamification, this work presents a cost-effective, engaging solution to foster physical activity, addressing both health and motivational barriers in sedentary populations.

\end{abstract}


\section{Introduction}

In recent years, the percentage of the population involved in office jobs increased, so sedentary-induced health problems become an occupational hazard. On the other hand, a lot of office positions involve overtime, which leaves workers with little time for regular gym visits.

Consequently, inactivity has turned into a widespread social issue. There are now a number of health issues that can be linked to inactivity. According to a survey on health problems conducted by the Chinese Center for Disease Control and Prevention, the annual incidence of low back pain among workers in China’s key industries is 16.4\% \cite{jia2022prevalence}, and working with the same posture with repetitive paces is a critical factor for such incidents.

Some companies are introducing radio calisthenics in the workplace \cite{zeng2024exercising}. However, the employees are reluctant to participate in such activities because of low motivation and lack of interest, as radio calisthenics are usually bonded with compulsory activities during school life. Therefore, in order to solve the problem of sedentary occupational hazards, the challenge is to make exercise fun with an easy implementation in an office environment, so that people can be motivated to exercise within their limited free time and working space.

The reasons for this include the lack of time for exercise and the fact that it is not interesting and therefore does not continue. In many Asian companies, radio calisthenics is a part of the daily exercise routine during work, and there are also opportunities for students to take part in radio calisthenics during class breaks and summer holidays, but some people find it uninteresting or unmotivating. Even for students at school who practice radio calisthenics, most people do the exercise arbitrarily without following the standard motions designed by the sports experts. As a result, radio calisthenics has failed to achieve the goal for exercises. In other words, the challenge is to make radio calisthenics fun and motivate people to do it, as well as giving people real-time feedback based on their performances via human-computer interaction, so that they can improve.

Currently, there are applications and wearable devices that promote exercise. These systems measure calorie expenditure and the travelling distance of the user, but their functions to motivate users to exercise are limited, and many have complicated settings during usage. Some somatosensory interactive games are also trying to solve this issue, such as games on Xbox and Nintendo platforms \cite{marks2015greater}. However, those devices are expensive and inappropriate to set up in an office environment.

In this study, we first used markerless posture estimation algorithms (OpenPose \cite{8765346} and PoseNet) to detect the keypoints. The method is proposed for detecting the joint positions of people when they practice radio calisthenics. We then calculate the joint angles in specific frames compared the standard joint angles in demonstration videos, to evaluate their performance and give them real-time feedbacks regarding their motions. Therefore, people can correct their motion according to the variance between their motions and the demonstraction. A lighter version of the motion detection algorithm is also deployed on mobile devices using PoseNet, so they can practice the radio calisthenics in outdoor environments.

To improve the fun part of the exercise, we also used fireworks, a familiar part of daily life and especially a festival tradition, as a subject. Fireworks are one of the familiar works of art that many people from various cultural backgrounds enjoy for celebrations \cite{liu2010japanese} \cite{cho2012predictors} \cite{maggs1976firework}. So, an exercise program with a firework theme will provide not only health benefits but emotional incentives to them as well. Based on the previous patterns of body motions to express particle movements \cite{xie2020body2particles} \cite{xie2020exercise}, we designed a system to detect the keypoints of the users' joint keypoints. Then the system will calculate the joint angles and compare the angles with the ones in the demonstration videos to guide the users' actions for correction and improvement and then we implemented the program to mobile devices that automatically generates various fireworks animations by adjusting parameters such as the angle, height, size, quantity, and shape of the fireworks used in the particle system in real-time based on the measured body movements.

This study focuses on the user's enjoyment of movement without being bound by predefined movement patterns or specific devices. The radio calisthenics training program associates the user's body movements with angles and rhythms to generate firework animations. With a computer vision model implemented on a mobile device, the fireworks are generated automatically in response to the user's body movements. The user can move his/her body in various ways to produce more beautiful fireworks, and it becomes a full-body exercise without notice.

\section{Related Works}

Existing exercise support systems often compare the user's posture with that of a model and correct the user's movements. For example, some Tai-chi training systems rely on VR or Openpose to present the difference between the practitioner's and the standardised model's motions \cite{iwaanaguchi2015cyber} \cite{qiao2017real}. Yao et al \cite{yao2019aproposal}, based on the locomotor exercises called `Active 5', aimed to make the exercises fun by determining the similarity between the user's postures and the model's in real time. Chan et al \cite{chan2010virtual} proposed a VR dance training program using motion capture technology to help students to learn dancing poses from a virtual teacher. The main focus of this type of work is to compare the user's motions to a standard model. They are helpful in sports practice but may demotivate the user to exercise efficiently. Moreover, most systems in this category require specific motion capture devices such as OptiTrack, Head Mounted Displays(HMD) or Kinect, which are not feasible to set up in a typical office or workplace environment.

\begin{figure}
    \centering
    \includegraphics[width=0.5\linewidth]{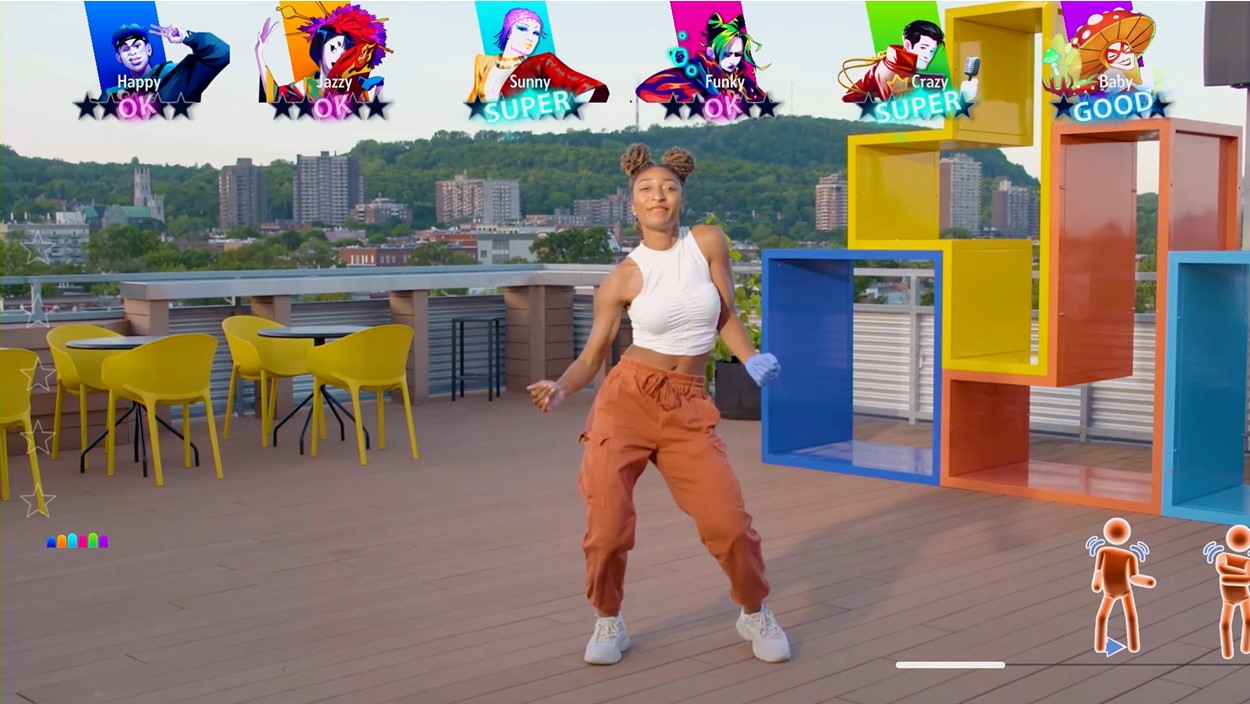}
    \caption{The screen cap of Just-Dance}
    \label{fig:justdance}
\end{figure}

\begin{figure}
    \centering
    \includegraphics[width=0.5\linewidth]{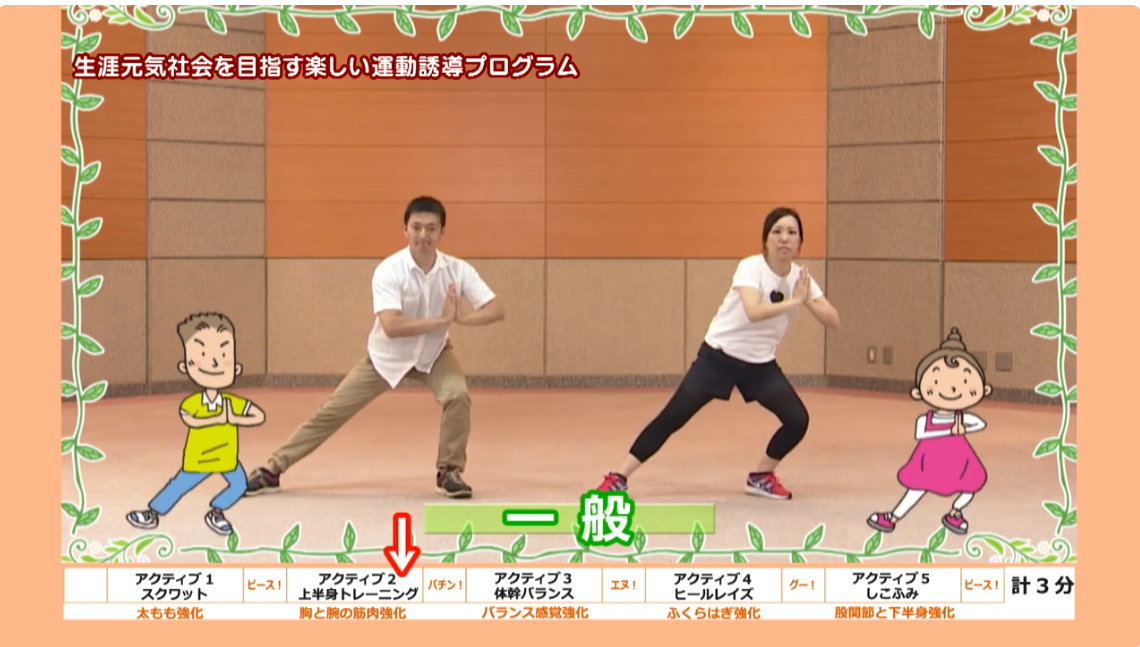}
    \caption{The screen cap of Active5}
    \label{fig:active5}
\end{figure}

Some game devices have physical interaction games. An interesting and fun exercise is `Just Dance' as shown in Fig \ref{fig:justdance}, a dance game in which the user dances with a person on the screen, through which scores and effects are generated. A special camera or external device is used to estimate the user's posture and calculate the score. However, the movements are complex, as they are performed in response to a song. Another fun, simple radio exercise is the `Active 5' \cite{ACTIVE5}, which is a locomotor exercise as shown in Fig \ref{fig:active5}. Based on the findings of health science, `Active 5' is an exercise that can be enjoyed by many generations at the same time, from children to the elderly, and can be easily tackled. It is characterised by `five exercises for lifelong vitality', `fun that three generations can synchronise together' and `step-up and a sense of achievement because it is not easy', and exercise can be performed easily in a short time. Also it requires a game device such as playstation or switch to work or even wearable devices, which are not feasible and proper to use in workplace and school environments.

Our work is mainly inspired by \cite{xie2020exercise} and \cite{Fireworks}. In their preliminary experiment, Xie et al \cite{xie2020exercise} recruited five subjects to use their bodies to represent the fireworks generated by a particle system in advance. Their body movements were recorded via a depth camera. Preliminary experiments showed that the subjects used their hands the most when describing the fireworks' launch, blooming, and shapes. The subjects used their upper limbs during the launching phase and used their hands to express the blooming of fireworks. They expressed the quantity and shape of the fireworks through the arms' opening angles. It was also found that the subjects tilted their bodies to the left and right to express the launch angle of the fireworks. This set of motions fits right in with the posture of the stretching exercise, and the motions are suitable for conducting in a workplace environment. However, their system also required Kinect as the motion capture device and created difficulties to deploy in an office environment, as the Kinect sensor is limited in the distance with a working range of between 1.2 and 3.5 meters. Also, the particle system they introduced in \cite{xie2020body2particles} required over twenty parameters based on a dense point cloud of the subject, so it is difficult to implement in a mobile device.

On the other hand, pose estimation methods based on sole RGB camera input have thrived in the past few years. We encourage the readers to refer to the comprehensive surveys as there are vast volumes of research papers published in this domain \cite{chen20232d, dubey2023comprehensive, zheng2023deep}. As many of the pose estimation methods are based on deep learning on workstations with GPUs \cite{8765346, yang2020unifying}, we decided to deploy a light pose estimation model, PoseNet \cite{PostureNet}, based on Tensorflow Lite \cite{david2021tensorflow} with RGB images as input so that it can be compatible with mobile devices.

Regarding the simulation of fireworks, the Python-based game development framework, PyGame \cite{Pygame}, provides a more efficient way to simulate fireworks in devices with limited computation power, such as a smartphone or a Raspberry Pi \cite{kelly2019python}. Multiple packages allow developers to design various fireworks shapes within the framework with limited parameters according to preset music rhythms \cite{Codegiovanni, Fireworks, youtubePlemaster012024}. Therefore, we utilised these works to simulate various shapes of fireworks in the gamification of the stretching exercise in an environment with limited space. 

\section{Method}

\subsection{Overall pipeline}

\begin{figure}
    \centering
    \includegraphics[width=1.2\linewidth]{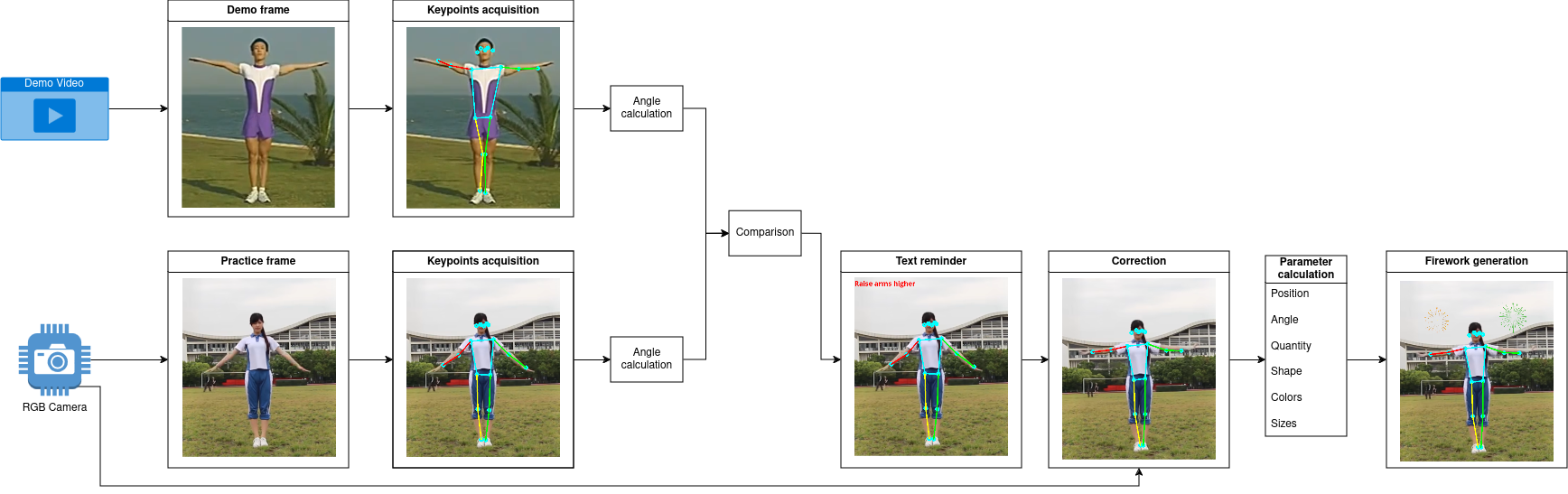}
    \caption{The pipeline of the system.}
    \label{fig:pipeline}
\end{figure}

The system configuration is shown in Figure \ref{fig:pipeline}. The user stands in front of the RGB-D camera and can simultaneously view his or her own image and the teacher's image. The system consists of three main parts: posture estimation, template matching, effect feedback, reward based on correction. Skeletal information is obtained from the user's real-time video and the teacher's demonstration video. For each skeletal information, template matching is performed to calculate the score of the current movement. Depending on the score, feedback of different effects appears in the real-time video. First, the system acquires joint data of the user's body from RGB images as input data. Next, it calculates the joint velocity, acceleration, and angle between joints to adjust the particle parameters in real-time. If the joint angles vary a lot from the teacher's actions and the user's actions, the system will directly output the text reminder on the screen for the user to improve their performance. If the user correct their motions, the system automatically generates different firework animations based on the body movements, as a reward of the correction. Figure \ref{fig:pipeline} shows the pipeline of the system.

We used the RGB image from the camera as an input. Then, the keypoints of the human body were extracted via either OpenPose or mobile PoseNet. We then calculated the parameters of the fireworks based on the user's motions. Finally, the fireworks are generated and shown on the screen.

\subsection{Human pose estimation}

Human pose estimation, also called human key points detection, is an essential task in computer vision. It is the preemptive task for human-computer interaction, human action recognition and behaviour analysis. Once the key points of a person can be recognized, we can use the key points to analyze their actions and design rewards and feedback based on the actions to close the loop in system design.

\begin{figure}
    \subfloat[Keypoints with their indices.]{%
        \resizebox*{6.5cm}{!}
        {\includegraphics{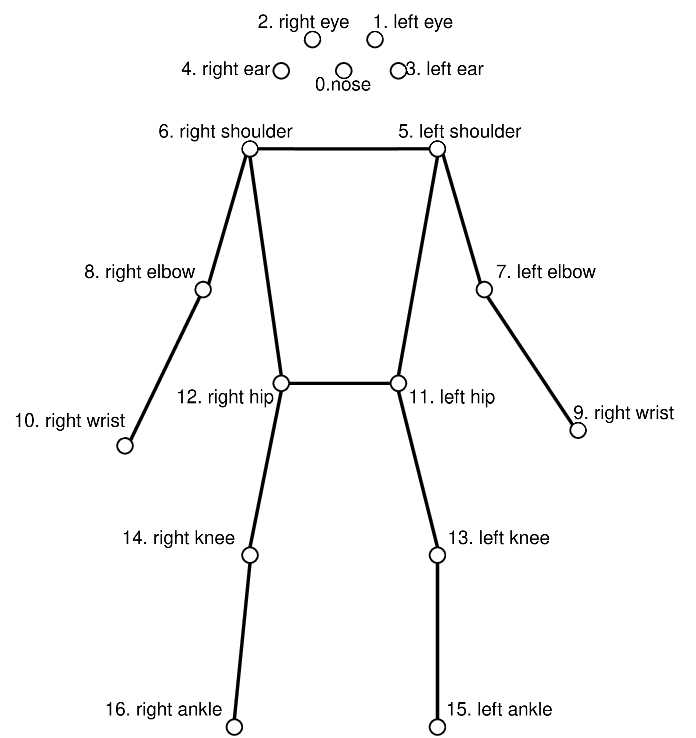}
        }
    }
    \hspace{5pt}
    \subfloat[Keypoints marked in a human body.]{%
        \resizebox*{6.5cm}{!}
        {\includegraphics{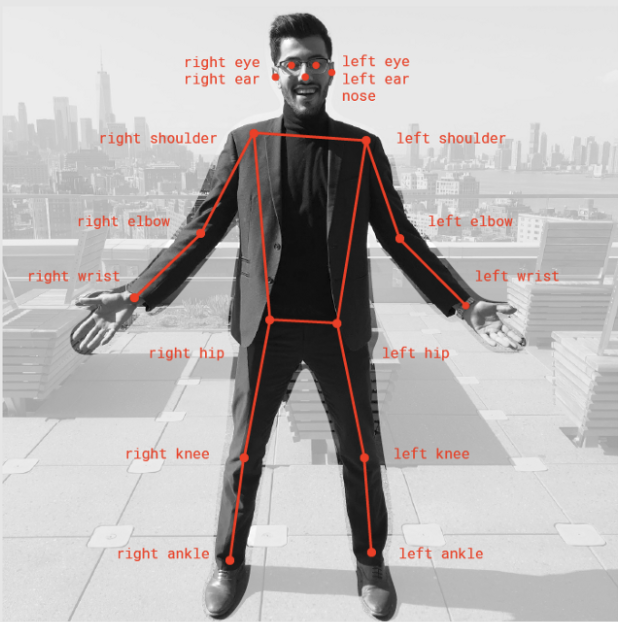}
        }
    }
\caption{Keypoints in COCO keypoints challenge.} \label{fig:keypoints}
\end{figure}

COCO keypoint challenge track is an authentic open competition in human pose estimation \cite{huang2020joint}. There are 17 joints marked as keypoint ground truth in the COCO dataset. The aim of human pose estimation is to detect the human and the relevant positions of the key points correspondingly. The key points are shown in Fig \ref{fig:keypoints}, as well as the corresponding positions on the human body \cite{agrawalPosture2021}. The left and right in the images are mirrored because of the optical structure of the camera.

\subsubsection{Openpose}

Human 2D posture estimate is the task of locating human body parts such as shoulders, elbows, and ankles in an input image or video. Most real-world applications of human pose estimation demand both high accuracy and "real-time" inference.

OpenPose \cite{8765346}, developed by Carnegie Mellon University researchers, might be regarded the cutting-edge technique to real-time human posture assessment. The code base is open-source on GitHub and quite well documented. Openpose was originally written in C++ and Caffe. As a method for extracting human poses in real time using deep learning, Openpose is capable of detecting multiple joint points on the body, face, and hands. It can analyze only camera images and videos without using special sensors such as accelerometers, and by using a high-performance processor such as a GPU, it can analyze images and videos in real time even when they contain multiple people.

The pipeline from OpenPose is actually quite basic and straightforward. First, an RGB input image is fed into a "two-branch multi-stage" CNN. Two branches indicate that the CNN produces two distinct outputs. Multi-stage simply refers to the network being stacked one on top of the other at each level. This step is equivalent to simply raising the neural network's depth in order to capture more refined outputs in later stages.

\begin{figure}
    \centering
    \includegraphics[width=\linewidth]{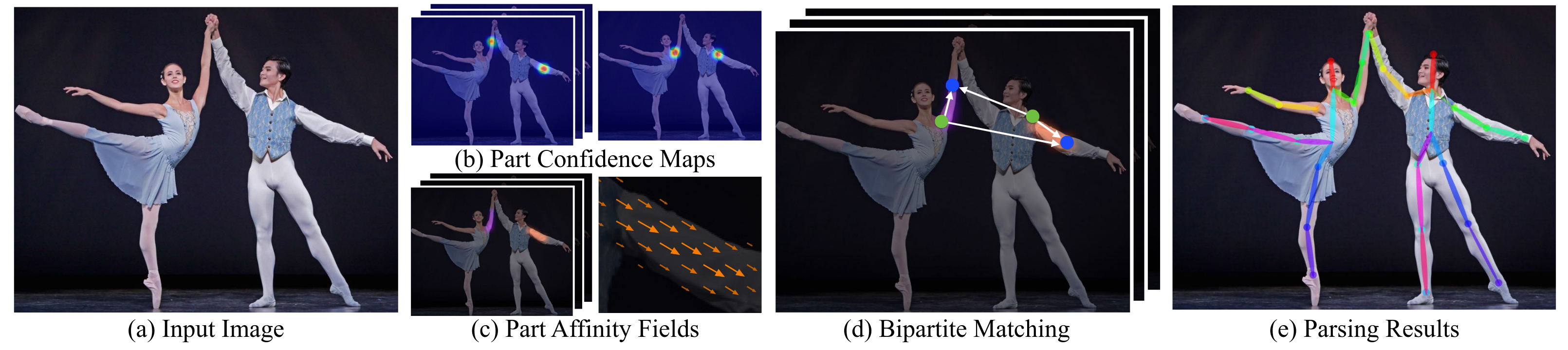}
    \caption{The pipeline of Openpose. Image taken from \cite{8765346}}
    \label{fig:openpose}
\end{figure}

First, an RGB input image (Fig \ref{fig:openpose}a) is fed into a "two-branch multi-stage" CNN. Two branches indicate that the CNN produces two distinct outputs. Multi-stage simply refers to the network being stacked one on top of the other at each level. This step is equivalent to just increasing the depth of the neural network in order to capture more refined outputs in later stages.

The data was then transmitted to a neural network with two branches. The top branch, depicted in beige, predicts the confidence maps (Fig \ref{fig:openpose}b) of multiple body parts' locations. The bottom branch, depicted in blue, predicts the affinity fields (Fig \ref{fig:openpose}c), which indicate the degree of relationship between distinct bodily components.

\begin{figure}
    \centering
    \includegraphics[width=\linewidth]{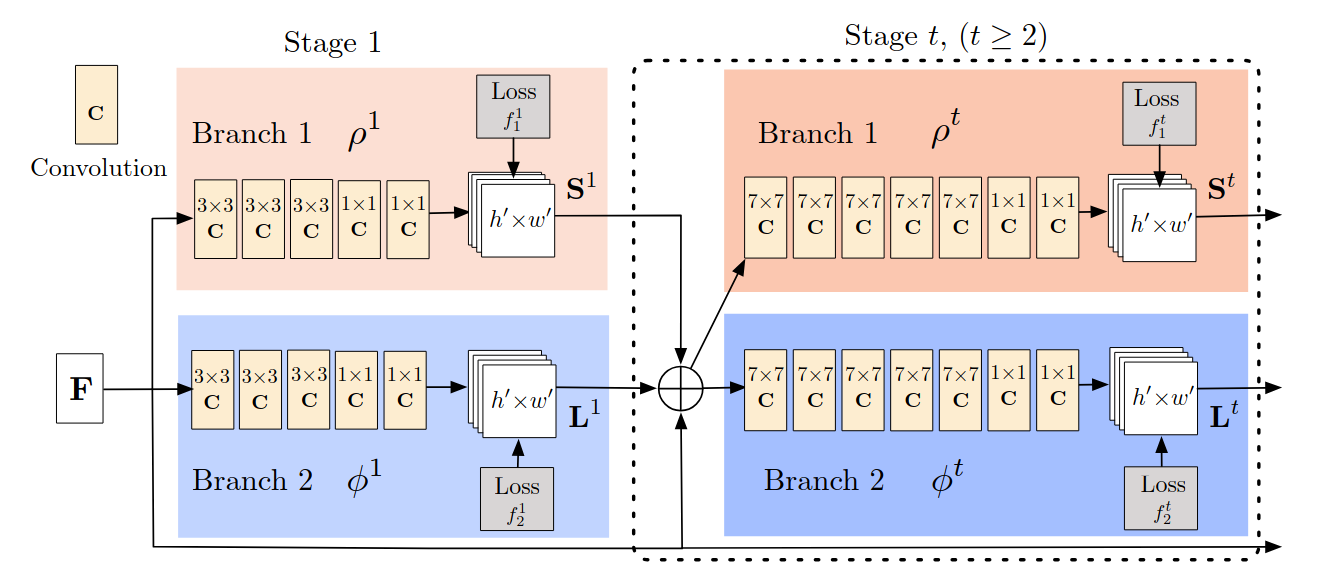}
    \caption{Architecture of the two-branch multi-stage CNN. Image taken from \cite{8765346}}
    \label{fig:openpose-arch}
\end{figure}

During the first stage (left half of Fig \ref{fig:openpose-arch}), the network generates an initial set of detection confidence maps $S$ and part affinity fields $L$. In following stages (right half of Fig \ref{fig:openpose-arch}), the predictions from both branches in the previous stage, as well as the original picture features $F$, are concatenated (shown by the $\oplus$ symbol in Fig \ref{fig:openpose-arch}) and utilized to construct more refined predictions. In the OpenPose implementation, the last stage $t$ is set to 6.

\begin{figure}
    \centering
    \includegraphics[width=\linewidth]{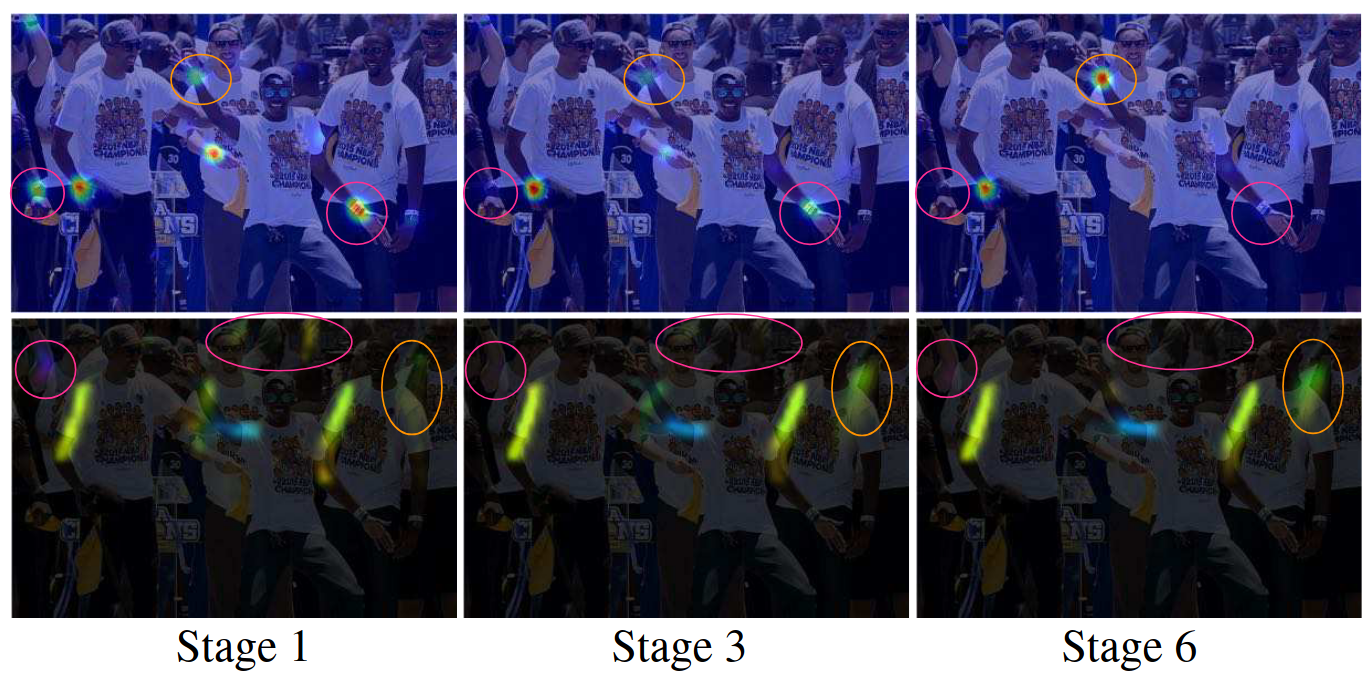}
    \caption{Outcome of a multi-stage network. The top row shows the network predicting confidence maps of the right wrist while the bottom row shows the network predicting the Part Affinity Fields of right forearm (right shoulder — right wrist) across different stages.}
    \label{fig:confidence_map}
\end{figure}

Figure \ref{fig:confidence_map} illustrates the advantages of a multi-stage configuration. In the example, we see that there is some confusion between the left and right body components in the early phases. However, as the stage advances, the network improves at making these distinctions.

Finally, greedy inference (Fig \ref{fig:openpose}d) processes the confidence maps and affinity fields to produce the 2D key points for all people in the image (Fig \ref{fig:openpose}e).

Referring back to Figure \ref{fig:openpose-arch}, the top branch of the neural network generates a set of detection confidence maps $S$. The mathematical definition is as follows:

\begin{equation}
    S = (S_1, S_2, \cdots, S_j), S_j \in \mathbb{R}^{w \times h}, j \in \{1, \cdots, J\} 
\end{equation}

where $J$ is the total number of body parts. $J$ is determined by the dataset used to train OpenPose. For the COCO dataset, $J = 18$ since there are $17$ separate body keypoints and $1$ backdrop. The picture below depicts the various bodily parts and their allocated ID for the COCO dataset.

For model trained with the COCO dataset, the set $S$ will have elements of $S_1, S_2, S_3, \cdots, S_{17}$ representing the confidence map of each body part . For this example, let’s assume that the element $S_1$ corresponds to the confidence map for the key point id of 0 (in Fig \ref{fig:keypoints}a) which refers to the noses. Therefore, only local areas in the image with high confidence in $S_1$ will be used to detect the noses.

Referring back to Figure \ref{fig:openpose-arch}, the neural network's bottom branch generates a set of part affinity field maps L. The mathematical definition is as follows. The total number of limbs, $C$, is determined by the dataset used to train OpenPose. To be clear, the paper refers to part pairings as limbs, even if some body part pairs are not human limbs. Each element in the set $L$ represents a map of size $w \times h$, with each cell containing a 2D vector indicating the direction of pair elements. For example, in Figure \ref{fig:openpose}, the body part pair extends from the right shoulder to the right elbow. The diagram then depicts a directed vector extending from the right shoulder to the right elbow.

The image is first evaluated by a pre-trained convolutional neural network, such as the top 10 layers of VGG-19, which generates a set of feature mappings F. The choice of feature extractor to make F is not restricted to VGG-19. Other variations of OpenPose use Mobilenet (which will be introducted in subsection \ref{posenet}) to extract visual features before transferring them to the rest of the neural network in \ref{fig:openpose-arch}.

At the first stage, the network generates a collection of detection confidence maps $S$ and part affinity fields $L$. The function variable $\rho$ represents the CNN that receives input $F$ and produces the output map $S$. The function variable $\phi$ represents the CNN that receives input $F$ and produces the output map $L$. The annotation "1" at the top of each symbol denotes inference in the initial step. In the following steps, the predictions from both branches in the previous stage, as well as the original image attributes $F$, are combined and used to generate more refined predictions. Therefore,

\begin{equation}
\begin{split}
    S^1 &= \rho^1(F) \\
    L^1 &= \phi^1(F) \\
    S^t &= \phi^t(F, S^{t-1}, L^{t-1}), \forall t \geq 2 \\
    L^t &= \phi^t(F, S^{t-1}, L^{t-1}), \forall t \geq 2
\end{split}
\end{equation}

To train the network how to generate the best sets of $S$ and $L$, Openpose applies two loss functions at the conclusion of each stage, one for each branch. The research applies a conventional $L2$ loss between calculated forecasts and ground truth maps and fields. Furthermore, weights are introduced to the loss functions to address the practical issue of some datasets failing to fully classify all people. The loss functions at a particular stage $t$ are shown below.

\begin{equation}
\begin{split}
    f^t_{S} &= \sum^{J}_{j=1} \sum_{p} W(p) \cdot ||S^t_{j}(p) - S^*_{j}(p) ||^2_{2}, \\
    f^t_{L} &= \sum^{C}_{c=1} \sum_{p} W(p) \cdot ||L^t_{c}(p) - L^*_{c}(p) ||^2_{2}
\end{split}
\end{equation}

The symbol $p$ refers to a single pixel point in a $w \times h$ picture. The $*$ notation next to the set $S$ and $L$ indicates that it is the ground truth. $S(p)$ returns a one-dimensional vector with the confidence score for the body part $j$ at picture point $p$. $L(p)$ returns a two-dimensional vector containing the directional vector for the limb $c$ at point $p$ in the image. $J$ denotes the total number of body components, while $C$ denotes the total number of "limbs" or body joint to body joint linkages. W(p) denotes the weighing function, as previously stated. When an image position $p$ lacks an annotation, $W(p)$ equals zero. The mask prevents true positive predictions from being penalized during training.

Finally, combining the two loss functions, the overall objective function of Openpose is $f = \sum^T_{t-1}(f^t_S + f^t_L)$.

In the preliminary experiment phase, Openpose provides accurate estimation for body joints so that we can calculate the angles between the joints. However, it relies on deep models like VGG-19 and Caffe, which is difficult to implement in mobile devices. So we moved to PoseNet to expand the training system to be more handy for end users.

\subsubsection{PoseNet}
\label{posenet}

As we would also like to deploy our system on a mobile device, we also chose the PoseNet based on Mobile Net for human pose estimation. Posenet is a real-time pose identification method that can determine individuals' positions in images or videos. It functions in both single-mode (detecting a single human posture) and multi-pose (detecting multiple human poses). Based on the TensorFlow platform, Posenet enables the developers to utilize a mobile device to estimate human posture by identifying body key points such as the elbows, hips, wrists, knees, and ankles. By connecting these points, the developer can create a skeletal structure of the human pose.

\begin{figure}
    \centering
    \includegraphics[width=\linewidth]{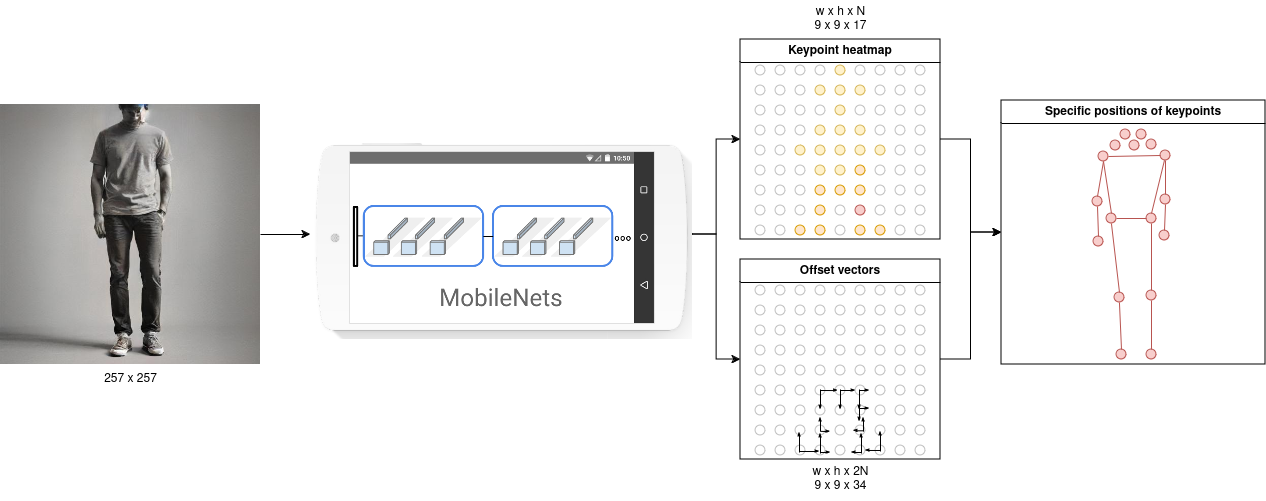}
    \caption{The overall framework of PoseNet}
    \label{fig:posenet-framework}
\end{figure}

The framework of our deployed version of PoseNet is shown in Fig. \ref{fig:posenet-framework}.  It was trained using the MobileNet Architecture, which optimized the classic convolution layer into a $1 \times 1 \times N$ deep convolution kernel and a single-layer complete matrix, making the convolution computation easier to handle on compuatation units of mobile devices. This lightweight model reduces parameters, increases accuracy, and deepens the network using depth-wise separable convolution.

The input of the network was $257 \times 257$ pixel images. After passing several mobile convolution layers, the input image was compressed into a $9 \times 9$ latent matrix. Then two heads were introduced as the keypoint heatmap and the offset vectors. The keypoint heatmap aimed to predict the rough position of specific key points, so its shape was $9 \times 9 \times 17$, where 17 was the number of the key points. The offset vectors aimed to calculate the $x$ and $y$ increments from the rough positions to the key points, so its shape was $9 \times 9 \times 34$, where $34$ was the number of the key points multipled by $2$, representing $x$ and $y$ directions.

At the end of human pose estimation stage, we got the relevant positions of keypoints of the user for motion analysis in the next stage.

\subsection{Angle calculation and comparison}

As the facial information is not important for radio calisthenics training, Head, Right Shoulder, Right Elbow, Right Wrist, Left Shoulder, Left Elbow, Left Wrist, Right Hip, Right Knee, Right Ankle, Left Hip, Left Knee, Left Ankle are the only 13 joints used in the angle calculation and comparison. In this study, the cosine similarity of two neighbouring limbs is computed. Initially, the limb vector $n_i$ is calculated using Equation below.

To compare the angle similarity of the user to the demo video, it is necessary to consider the time difference between the postures of the two individuals when the chameleon effect occurs and the posture information of the entire time series of each of the two individuals to be judged. In this study, OpenPose is used to acquire the posture information. Regarding the time difference between the two postures when the chameleon effect occurs, the study by Sparenberg et al. \cite{sparenberg2012minimal} suggests that a time difference of 2-4 seconds between the two postures when the chameleon effect occurs is appropriate, and other studies have conducted experiments assuming a delay time of 2-4 seconds. Therefore, in this study, a 3-second delay is taken into comparison.

\begin{equation}\label{eq:vector}
    n_i = (X_i, Y_i) - (X_{i+1}, Y_{i+1}), 0 \leq i \leq 12
\end{equation}

Equation \ref{eq:vector} gives the vector set $P = (n_1, n_2, . . . n_{12})$ for the 12 limbs. We then calculate the cosine values between the vectors of the two neighbouring limbs $n_i$ and $n_j$, calculated from Equation 2 below.

\begin{equation}\label{eq:cos}
    \cos \alpha_i = \frac{\sum^n_{i=1}\sum^n_{j=1}(n_{i} \times n_j)}{\sqrt{\sum^n_{i=1}(n_i)^2 \times \sum^n_{j=1}(n_j)^2}}
\end{equation}

The user's joint data and the demonstration video teacher's joint data are combined in Equations \ref{eq:vector} and \ref{eq:cos} with their respective cosine values $\theta_{\text{demo}} = \{\alpha_{d_1}, \alpha_{d_2}, \cdots, \alpha_{d_{12}}\}$ and $\theta_\text{practice} =  \{\alpha_{p_1}, \alpha_{p_2}, \cdots, \alpha_{p_{12}}\}$ are calculated. Then we calculate the sum of the difference values of these two cosine values.

\begin{equation}
    \alpha_\text{sum} = \sum^{12}_{i=1} (\alpha_{d_i} - \alpha_{p_i})
\end{equation}

However, there are times during the radio calisthenics when some joints do not move very much. To reduce the influence of such joints, weights $w_i$ according to Equation \ref{eq:weight} are given to the joints that move well. Each of the 12 joints is multiplied by its weight, giving a total $D$.

\begin{equation}\label{eq:weight}
\begin{split}
     w_i &= \frac{1-e^{-\alpha_i/\alpha_\text{sum}}}{\sum^{12}_{i=1}(1-e^{-\alpha_i/\alpha_\text{sum}})} \\
     D &= \sum^{12}_{i=1} \alpha_i \cdot w_i
\end{split}
\end{equation}

Finally, a score of similarity is calculated as in Equation \ref{eq:score} below. In this study, $D_\text{std}$ was given a standard deviation value of 65.

\begin{equation}\label{eq:score}
  S =
    \begin{cases}
      (D_{\text{std}}-D) \cdot \frac{100-S_\text{std}}{D_\text{std}} + S_\text{std} & 0 \leq D \leq D_\text{std} \\
      0 & D > D_\text{std}
    \end{cases}
\end{equation}

\begin{figure}
    \subfloat[Demonstration motion]{%
        \resizebox*{5cm}{!}
        {\includegraphics{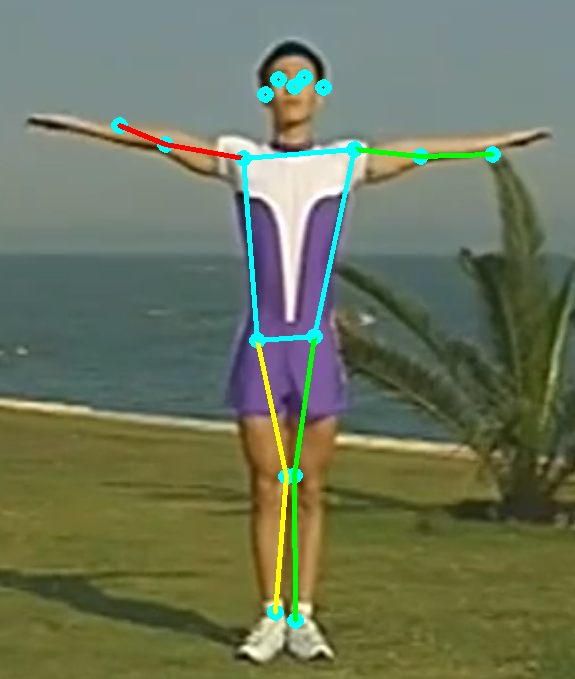}
        }
    }
    \subfloat[Practice motion]{%
        \resizebox*{5cm}{!}
        {\includegraphics{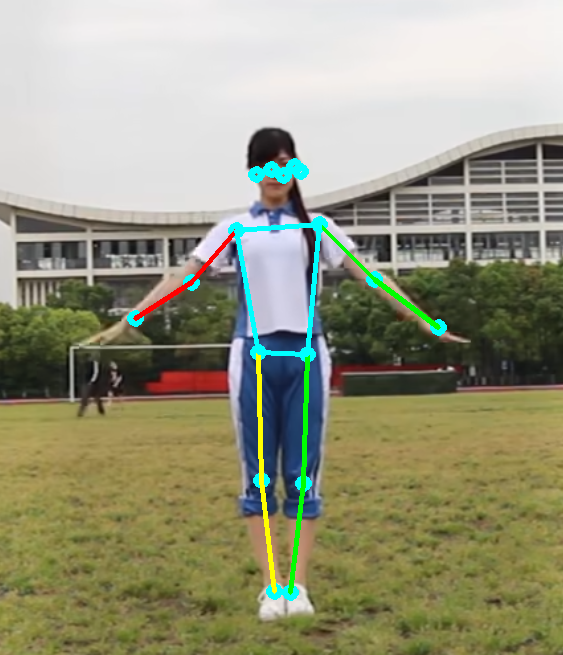}
        }
    }
    \subfloat[Reminder]{%
        \resizebox*{5cm}{!}
        {\includegraphics{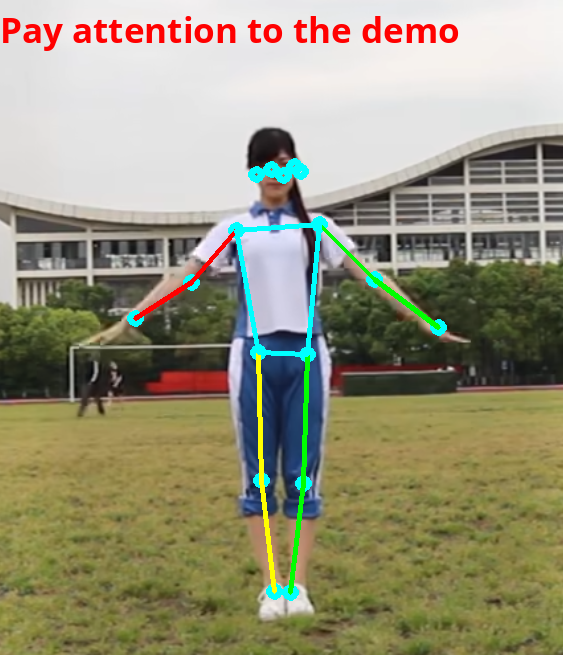}
        }
    }
\caption{Reminder for the user.} \label{fig:reminder}
\end{figure}

If the score is 0, the system will raise a reminder to the user on the screen so that they can pay attention to the demonstration video. As show in Figure \ref{fig:reminder}. Since the user raised the arms too low compared to the demonstration video, the score is lower than the threshold and therefore the reminder is raised for the user to correct the action.

\subsection{Motion Correction and parameter calculation for reward}

If the user correct the motion and the score increased above zero, then the system will generate fireworks as a reward to the user.

Based on the results of preliminary experiments in \cite{xie2020exercise}, we linked the parameters of the fireworks that is linked to body movements. The mapping between the parameters of the fireworks and the user's body movements is introduced below.

\begin{figure}
    \centering
    \subfloat[Star shape]{%
        \resizebox*{4.5cm}{!}
        {\includegraphics[trim={0 2.3cm 0 0},clip]{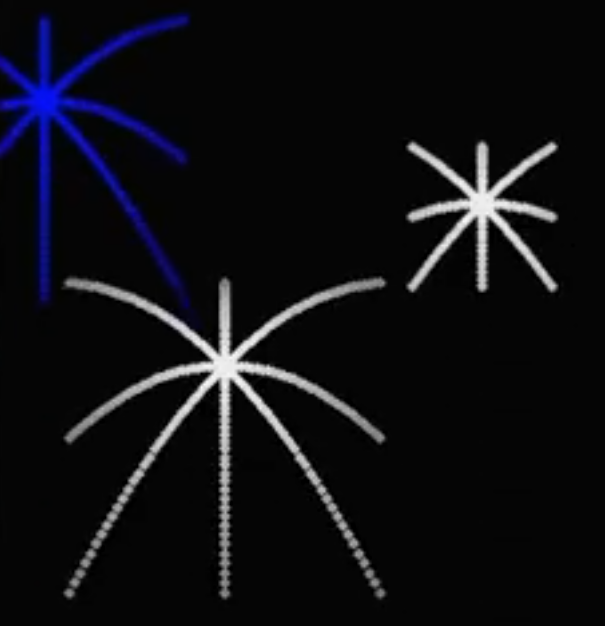}
        }
    }
    \hspace{1pt}
    \subfloat[Ball shape]{%
        \resizebox*{4.5cm}{!}
        {\includegraphics{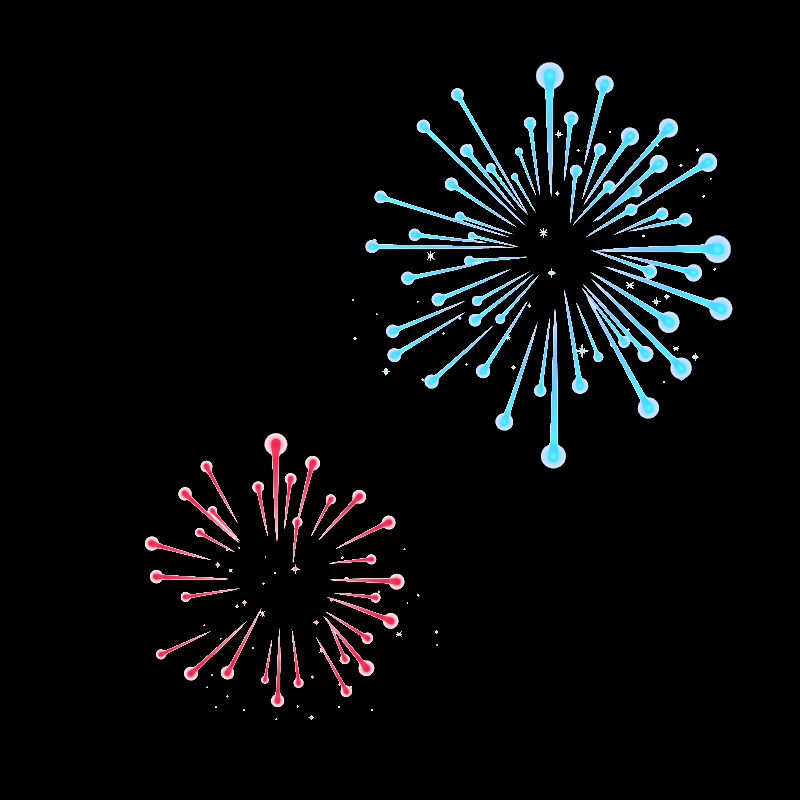}
        }
    }
    \hspace{1pt}
    \subfloat[Cluster shape]{%
        \resizebox*{4.5cm}{!}
        {\includegraphics[trim={0 1cm 0 0},clip]{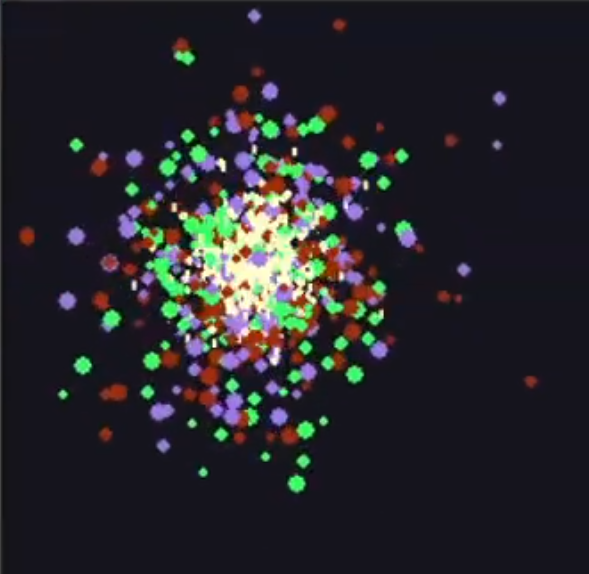}
        }
    }
\caption{Different shapes of fireworks} \label{fig:fireworks}
\end{figure}

\begin{itemize}
    \item The positions $x$ and $y$ of the fireworks: the positions of the firweorks were based on the activity of the body parts as shown in Equation (1):
    \begin{equation}
    \begin{aligned}
        x_{\text{firework}} &=\{  {X_{i}, \text{ if } \|X_{i_{t}} - X_{i_{t-1}\| > \text{threshold} }} \} \\
        y_{\text{firework}} &= \{  {Y_{i}, \text{ if } \|Y_{i_{t}} - Y_{i_{t-1}\| > \text{threshold} }} \} \\   
    \end{aligned}
    \end{equation}
    \item The quantities of fireworks: The quantity of the fireworks depends on the number of active joints in consecutive frames. If multiple active joints were detected in two consecutive frames, multiple fireworks will be generated.
    \item The angle $\theta$ of the fireworks: 90 $^{\circ}$ when the user is standing upright, and calculating from Equation (2) based on the shoulder and hip coordinates:
    \begin{equation}
        \theta = \arctan \frac{X_{\text{left shoulder}} + X_{\text{right shoulder}} - X_{\text{left hip}} - X_{\text{right hip}}}{Y_{\text{left shoulder}} + Y_{\text{right shoulder}} - Y_{\text{left hip}} - Y_{\text{right hip}}}
    \end{equation}
    \item The shapes of the fireworks: We designed three types of firework shapes as shown in Fig \ref{fig:fireworks}. The types of fireworks to be generated depend on the movement amplitudes of the active joints. If the amplitude is small, a star-shaped firework will be launched. If the amplitude is medium, an ball-shaped firework will be launched. Considering that neck and shoulder stiffness is a typical problem in office ergonomics, a cluster-shape firework will be launched if the user raises their hands over the head.
    \item The colors of the fireworks: Regarding color adjustment, the average heights of multiple joints including nose, shoulders, elbows, hips, knees and wrists are used to perform a six-step color adjustment using Equation (3).
    \begin{equation}
    \begin{aligned}
  \text{color}=\left\{
    \begin{array}{ll}
      \text{white}, & \mbox{if $\overline{Y}_{\text{knees}} > \overline{Y}_{\text{wrists}}$}.\\
      \text{purple}, & \mbox{if $\overline{Y}_{\text{hips}} > \overline{Y}_{\text{wrists}} > \overline{Y}_{\text{knees}}$}.\\
      \text{blue}, & \mbox{if $\overline{Y}_{\text{hips}} > \overline{Y}_{\text{wrists}} > \overline{Y}_{\text{knees}}$}.\\
      \text{green}, & \mbox{if $\overline{Y}_{\text{elbows}} > \overline{Y}_{\text{wrists}} > \overline{Y}_{\text{hips}}$}.\\
      \text{orange}, & \mbox{if $\overline{Y}_{\text{shoulders}} > \overline{Y}_{\text{wrists}} > \overline{Y}_{\text{elbows}}$}.\\
      \text{yellow}, & \mbox{if $\overline{Y}_{\text{wrists}} > Y_{\text{head}}$ and $\overline{Y}_{\text{elbows}} < Y_{\text{nose}}$.} \\ 
      \text{multiple colors}, & \mbox{if $\overline{Y}_{\text{wrists}} > Y_{\text{head}}$ and $\overline{Y}_{\text{elbows}} > Y_{\text{nose}}$.}
    \end{array}
  \right.
    \end{aligned}
    \end{equation}
    \item The sizes of the fireworks: The size of the firework after blooming was set at four levels, and is determined by Equation (4). As shown in the equation, the maximum horizontal distance $H_{\text{max}}$ was determined by the maximum of the distances between the left and right wrists, left and right elbows, left and right ankles, and left and right shoulders at the time of launching.
    \begin{equation}
    \begin{aligned}
    \text{size}=\left\{
    \begin{array}{ll}
      \text{large}, & \mbox{if $H_{\text{max}} = \|X_{\text{left wrist}} - X_{\text{right wrist}}\|$}.\\
      \text{medium}, & \mbox{if $H_{\text{max}} = \|X_{\text{left ankle}} - X_{\text{right ankle}}\|$}.\\
      \text{small}, & \mbox{if $H_{\text{max}} = \|X_{\text{left elbow}} - X_{\text{right elbow}}\|$}.\\
      \text{tiny}, & \mbox{if $H_{\text{max}} = \|X_{\text{left shoulder}} - X_{\text{right shoulder}}\|$}.
    \end{array}
  \right.
    \end{aligned}
    \end{equation}
\end{itemize}

\begin{figure}
    \centering
    \subfloat[The correction motion made by the user.]{%
        \resizebox*{5cm}{!}
        {\includegraphics{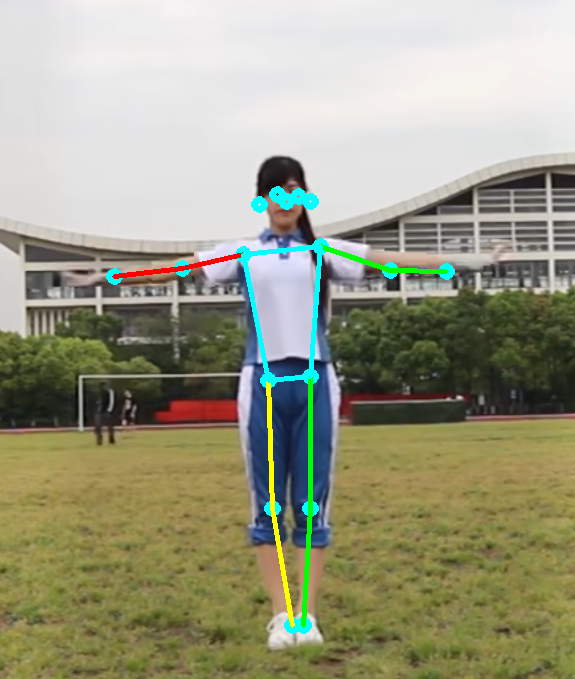}
        }
    }
    \hspace{5pt}
    \subfloat[Fireworks generated by the motions.]{%
        \resizebox*{5cm}{!}
        {\includegraphics{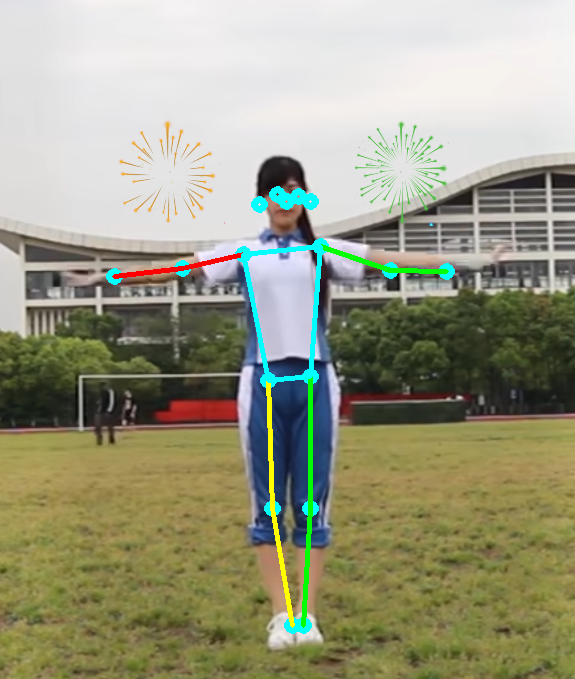}
        }
    }
\caption{Fireworks generated based on two frames.} \label{fig:frames}
\end{figure}

A generated firework image based on two consecutive frames is shown in Fig. \ref{fig:frames} . As two medium correct motions of the elbows and wrists are detected, two ball-shape fireworks fireworks would be launched. As at one side, the height of the elbow were between the shoulders and the elbows and on the other side, the height of the elbow is higher than the wrist and the wrist is above the hip, the color of one fireworks is green and the other will be orange. As the maximum horizontal distance pair was left and right wrists, the size of the fireworks would be large.

\section{Experiments and user experience evaluation}

\subsection{Experiments}

This experiment aims to test the following hypotheses:

1.Effectiveness: The real-time feedback provided by the system can significantly reduce the Angle error of user movements and improve the standardization degree of broadcast gymnastics.

2.Fun: fireworks generation mechanism can effectively improve user interest in participation and long-term use intention.

3.Universality: The system has applicability in different age and occupation groups, and it is feasible to deploy on mobile terminals.

\begin{figure}[!ht]
    \centering
    \includegraphics[width=0.5\linewidth]{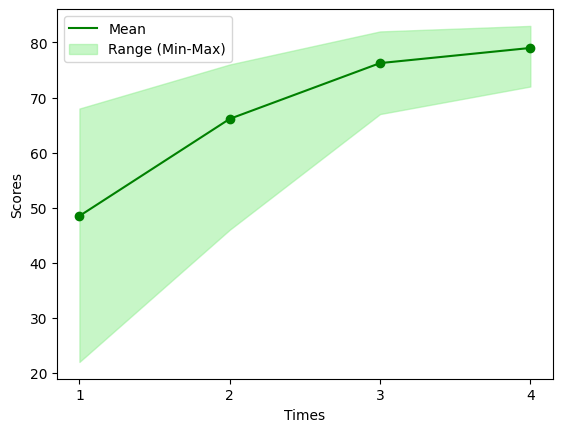}
    \caption{The scores of the participants in four experiments by using the system.}
    \label{fig:score}
\end{figure}

We used 136 people from Southwest Jiaotong University, including college students, cleaning workers, teachers, dormitory administrators and other evenly distributed in various age groups(There were 68 men and 68 women) participated in a series of experiments involving radio calisthenics, guided by a demonstration video provided by the system. Each participant performed the exercise four times, with their individual scores capped using a ceiling operation for statistical purposes. Table \ref{tab:score} shows the mean, maximum, and minimum individual scores across the four sessions. And Figure \ref{fig:score} showed the trend of improvement during the experiment.

\begin{center}
\begin{tabular}{||c c c c||} 
 \hline
Experiment & Mean Score & Maximum Score & Mininum Score \\ [0.5ex] 
 \hline\hline
 1 & 48.5 & 68 & 22 \\ 
 \hline
 2 & 66.13 & 76 & 46 \\
 \hline
 3 & 76.25 & 82 & 67 \\
 \hline
 4 & 81 & 83 & 72 \\
 \hline
\end{tabular}
\label{tab:score}
\end{center}

In Experiment 1, the participants' performance was initially assessed with an average score of 48.5.  This was accompanied by a maximum score of 68, indicating some variation in individual performance, and a minimum score of 22, which suggests that a few participants struggled significantly during the task.  However, by the time Experiment 2 came around, there was a noticeable improvement in the participants' overall performance.  The mean score increased to 66.13, reflecting a marked enhancement in the group’s ability to execute the tasks.  The highest score also improved, rising to 76, while the lowest score saw an increase to 46, suggesting that even participants who had previously struggled showed progress.

This upward trend continued into Experiment 3, where the participants’ mean score climbed further to 76.25.  At this stage, the range of scores became narrower, with the maximum score reaching 82 and the minimum score improving to 67.  This indicates that not only did the overall performance improve, but there was also a more consistent level of achievement across participants, with fewer outliers at the lower end of the scale.

By the final stage of the experiment, Experiment 4, the average score had risen to 81, with the highest score reaching 83 and the lowest score at 72.  These results show that the majority of participants had continued to improve, with the performance gap between the highest and lowest scores narrowing even further.

The consistent increase in scores across all four experiments strongly suggests that repeated practice with the system's demonstration video played a significant role in enhancing participants' performance.  Particularly noteworthy is the improvement seen in participants who initially had low scores.  This can be attributed to the real-time feedback provided by the system, which allowed participants to quickly identify when their movements deviated from the demonstration.  This immediate feedback enabled them to focus on correcting their weaknesses and to engage in continuous, targeted learning, which likely facilitated their improvement over time.  Thus, the system’s interactive nature and its ability to offer personalized guidance appear to have contributed significantly to the overall progress observed in the study.

\subsection{User experience evaluation}

In addition to participating in the exercise, 136 participants were interviewed and rated the following four aspects on a 5-point scale:

\begin{itemize}
	\item The entertainment value of creating fireworks with body movements.
	\item The ability to create different types of fireworks through various body motions.
	\item Whether the system effectively promotes exercise.
	\item The user-friendliness of the system on mobile devices.
\end{itemize}

As shown in Fig. \ref{fig:chart}, the results of the questionnaire indicated that all participants considered the system excellent in encouraging exercise, with overall ratings being high. For instance, all participants found it fascinating to create fireworks through body motions. Several participants shared comments such as: "It's captivating that body movements can generate different fireworks animations," "Many of the firework patterns are incredible," and "The system definitely encourages people to be more physically active."

In terms of user experience, the feedback revealed several strengths of the system: most users felt it effectively encouraged exercise, especially due to the interactivity of creating fireworks. Furthermore, participants generally found the system’s interface on mobile devices intuitive and easy to use, which contributed to a more immersive experience. Of course, this is part of the questionnaire feedback.

\subsection*{User A (University Student)}
\textbf{Background:} A 22-year-old student with a long history of sitting while studying, leading to a hunchback. The initial angle error was 28° (shoulder forward tilt). \\
\textbf{System Intervention:} The "firework height-spine extension" linkage mechanism was applied. When the back was straightened, golden fireworks were generated. \\
\textbf{Result:} After 3 weeks of practice, the angle error reduced to 9°, and the daily practice time increased from 5 minutes to 15 minutes. \\
\textbf{Feedback:} "The golden fireworks felt like unlocking an achievement, which made me more motivated to sit up straight while studying. My roommates even said my posture improved."

\subsection*{User B (University Lecturer)}
\textbf{Background:} A 35-year-old fitness coach who seeks precise movement control. The initial error was 18° (elbow abduction). \\
\textbf{System Adaptation:} The "Expert Mode" was activated, providing joint angle feedback and advanced firework combinations (e.g., spiral trajectories). \\
\textbf{Result:} The error was optimized to 5°, and the user designed seven custom firework patterns. \\
\textbf{Feedback:} "The data-driven feedback allows me to fine-tune my movements. I now use this system to demonstrate correct posture while training my clients."

\subsection*{User C (Retired University Professor)}
\textbf{Background:} A 68-year-old recovering from a rotator cuff injury, with limited shoulder flexion (maximum 45°). The initial error was 25°. \\
\textbf{System Adjustment:} The system automatically detected injury constraints and reduced the effective movement range to 30°-60°, generating gentle "starlight particle" effects. \\
\textbf{Result:} After 8 weeks, the shoulder flexion improved to 75°, and the error decreased to 12°. \\
\textbf{Feedback:} "The system didn't force me into difficult movements. The starlight effects made my rehabilitation process feel full of hope."

\subsection*{User D (Programmer)}
\textbf{Background:} A 38-year-old university programmer who frequently sits while using a computer, leading to lower back strain. The initial lumbar rotation error was 34°. \\
\textbf{System Functionality:} The system used AR projection to map firework effects onto the wall, encouraging large-range lumbar movements. \\
\textbf{Result:} The error decreased to 14 percent, and the weekly practice frequency increased from 2 sessions to 6. \\
\textbf{Feedback:} "The AR fireworks turned my small apartment into a fitness stage. I could work and exercise at the same time, which significantly alleviated my back pain."\\

However, the system did face some challenges during usage. First, due to the limited accuracy of MobileNet for pose recognition, some participants experienced misidentification or even a complete failure to recognize their movements. This lack of accuracy affected the fluidity of the feedback, leading to a suboptimal user experience. Secondly, some users reported a delay between their body movements and the fireworks’ generation, as the system produces fireworks based on the image input. This time lag was particularly noticeable during fast movements and impacted the interactive perception of the system.

\begin{figure}
	\centering
	\includegraphics[width=\linewidth]{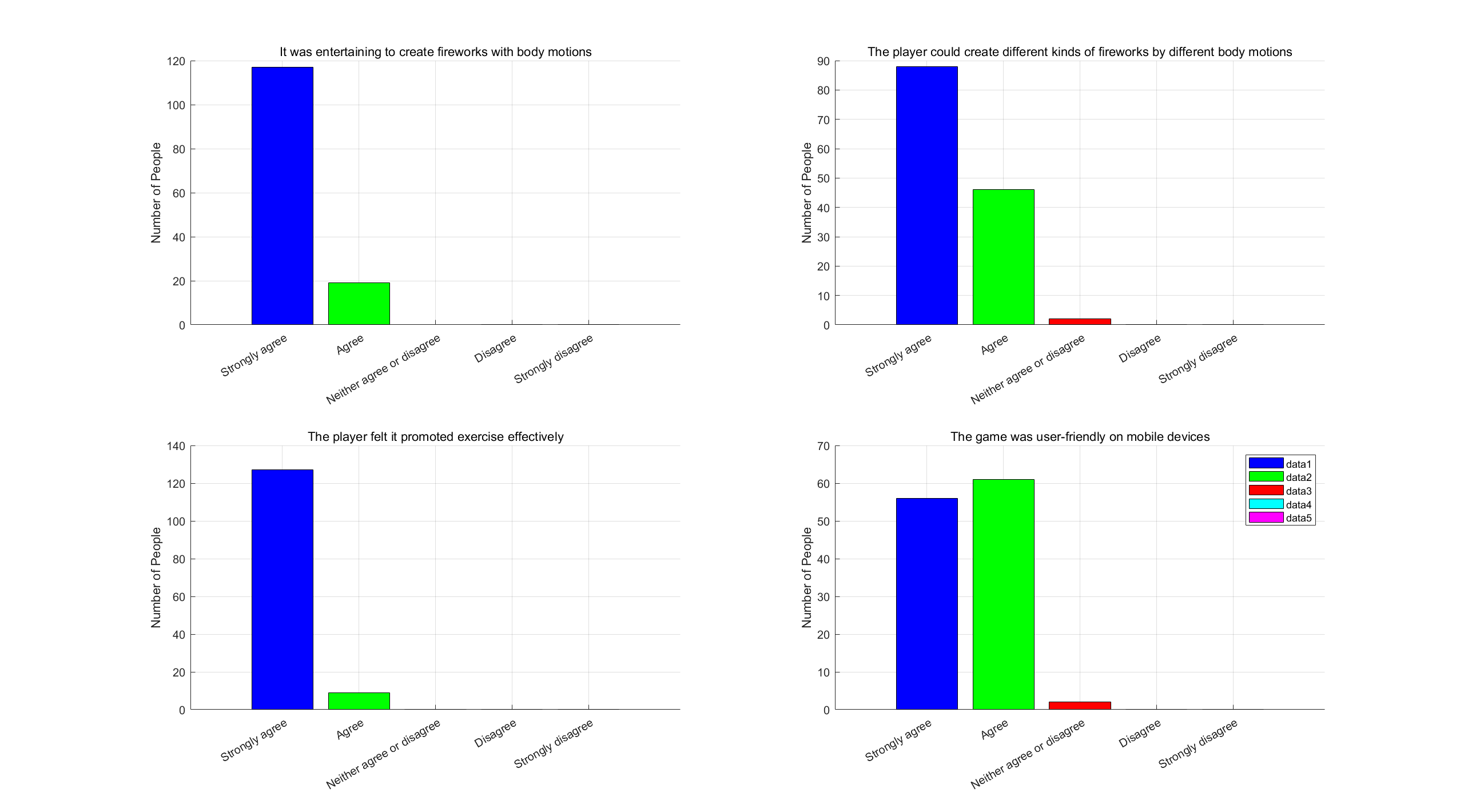}
	\caption{The evaluation of the system from the participates}
	\label{fig:chart}
\end{figure}\

As seen in Fig \ref{fig:chart}, the questionnaire's results indicated that every participant thought the system did an excellent job of encouraging exercise. Furthermore, we received remarks like "It's fascinating that body movements can produce different fireworks animations," "Many of the patterns are incredible," and "It encourages physical activity in people."

\textbf{"It was entertaining to create fireworks with body motions"}:
\begin{itemize}
	\item 117 users (85.4\%) \textbf{strongly agreed} with this statement.
	\item 19 users (13.9\%) \textbf{agreed}.
	\item No users selected "Neither agree nor disagree," "Disagree," or "Strongly disagree."
\end{itemize}

\textbf{"The player could create different kinds of fireworks by different body motions"}:
\begin{itemize}
	\item 88 users (63.8\%) \textbf{strongly agreed}.
	\item 46 users (33.3\%) \textbf{agreed}.
	\item 2 users (1.4\%) selected \textbf{"Neither agree nor disagree"}.
	\item Very few users selected "Disagree" or "Strongly disagree," making their responses negligible.
\end{itemize}

\textbf{"The player felt it promoted exercise effectively"}:
\begin{itemize}
	\item 127 users (93.4\%) \textbf{strongly agreed}.
	\item 9 users (6.6\%) \textbf{agreed}.
	\item No users selected "Neither agree nor disagree," "Disagree," or "Strongly disagree."
\end{itemize}

\textbf{"The game was user-friendly on mobile devices"}:
\begin{itemize}
	\item 56 users (40.6\%) \textbf{strongly agreed}.
	\item 61 users (44.2\%) \textbf{agreed}.
	\item 2 users (1.4\%) selected \textbf{"Neither agree nor disagree"}.
	\item 17 users (12.3\%) \textbf{disagreed}.
	\item No users selected "Strongly disagree."
\end{itemize}

The survey results indicate that most users highly appreciated the system's entertainment value, interaction effectiveness, and ability to promote physical activity. Among these, the statement \textbf{"The player felt it promoted exercise effectively"} received the highest level of agreement, with 93.4\% of users strongly agreeing. This suggests that the system successfully fulfills its core objective of encouraging physical movement.

However, regarding \textbf{"The game was user-friendly on mobile devices"}, while 84.8\% of users gave positive feedback (\textbf{strongly agree + agree}), 12.3\% of users expressed dissatisfaction. This may be attributed to factors such as limited visual experience on smartphones and constraints in interaction accuracy.

To enhance the system's performance and usability, the following improvements are proposed:
\begin{itemize}
	\item \textbf{Improve Motion Recognition Accuracy}: Implement advanced deep learning models, such as \textbf{EfficientNet} or \textbf{Vision Transformer}, to reduce misclassification and enhance interaction quality.
	\item \textbf{Optimize Interaction Latency}: Improve image processing and model inference efficiency to minimize delays between body movements and fireworks generation, increasing immersion.
	\item \textbf{Enhance User Guidance}: Introduce more detailed tutorials or real-time feedback mechanisms to help users quickly understand system functionalities.
	\item \textbf{Optimize Mobile Experience}: Adapt visual effects for smartphones or provide a PC-optimized version to accommodate different user needs.
\end{itemize}
These enhancements aim to further improve user experience, making the system more practical and widely acceptable.

\section{Conclusion and future work}

In this work, we proposed a motion-sensing system for mobile devices that is based on human pose estimation in order to encourage users to exercise and make exercise enjoyable. The user is free to shoot off fireworks by using their entire body. We verified that this system can help the users to focus on the motions they need to improve so that their motions can be more close to the demonstration motions by the teacher. The users eventually performed more standard motions in the radio calisthenics exercise. Also, there is a component of user discovery as we experienced the different fireworks that can be produced in tandem with the body movements.

In the future, we would like to design individual scores based on specific joint and limbs so that the reminder to the user can be more targeted. For example, guide the user to raise their arms higher or kick higher for specific motions. Also, there are new pose estimation methods to extract 3D poses based on 2D images but they are still heavy to deploy on mobile devices. If lighter models can be developed, we can analyze more motion patterns can give more specific guidance for the users regarding complex motions, such as dancing or martial arts.

We hope to expand the variety of fireworks that can be produced in the future and enhance the system's enjoyment. In addition, we would like to add music, a screen projection feature, and the ability for many users to compete against one another. We intend to increase the precision of motion recognition by other on-device networks in order to avoid motions being misidentified.

\bibliographystyle{plainnat}
\bibliography{main}


\section{Ethics and Morality Statement}

Informed Consent: All participants in this study provided informed consent prior to data collection. We ensured that participants fully understood the nature, purpose, and potential impact of their involvement and participated voluntarily.

Privacy Protection: All survey data is kept confidential. To safeguard participants' privacy, all personal information has been anonymized, and no identifying details of participants will be disclosed. The data will be used solely for the purposes of this research and will not be shared with any third parties.

Data Security: We have implemented necessary data protection measures, including data encryption and access control, to ensure the security and integrity of the data.

Ethics Review: This study has been approved by the Ethics Review Board (or relevant ethics committee) of our institution, ensuring that the research design complies with ethical standards and does not cause unnecessary psychological or physical harm to participants.

Transparency and Fairness: During the data analysis, we have ensured transparency in our research methods, and the findings have been reported impartially. All data analysis methods and results are truthfully presented, with no falsification, alteration, or concealment of data.

Accountability Statement: We commit to adhering to international ethical standards and respecting the rights and interests of participants. Any violations of ethics in the research process will be met with appropriate accountability.

This study is dedicated to ensuring the ethical compliance of all research activities and to treating participants with fairness, respect, and dignity throughout the process.

\end{document}